\pdfoutput=1

\documentclass[11pt]{article}

\usepackage[preprint]{acl}

\usepackage{times}
\usepackage{latexsym}

\usepackage[T1]{fontenc}

\usepackage[utf8]{inputenc}

\usepackage{microtype}

\usepackage{inconsolata}
\usepackage{booktabs}
\usepackage{multirow}
\usepackage{hyperref}
\usepackage{rotating}
\usepackage{enumitem}
\usepackage{tablefootnote}
\usepackage{pifont}

\usepackage{pdfpages}
\usepackage{longtable}
\usepackage{setspace}
\usepackage{etoolbox,xspace}
\usepackage{graphicx}
\usepackage{placeins}
\usepackage[toc,page]{appendix}

\usepackage{caption}
\usepackage{enumitem}
\usepackage{graphicx}
\usepackage{amssymb}
\usepackage{xspace}
\usepackage[ethiop,main=english]{babel}
\usepackage{inconsolata}
\usepackage[most]{tcolorbox}
\usepackage{graphicx}
\usepackage{pifont}
\usepackage{float}

\title{Whispering in Amharic: Fine-tuning Whisper for Low-resource Language}

\author{
 \textbf{Dawit Ketema Gete\textsuperscript{1}},
 \textbf{Bedru Yimam Ahmed\textsuperscript{1}},
 \textbf{Tadesse Destaw Belay\textsuperscript{1}},
 \\
 \textbf{Yohannes Ayana Ejigu\textsuperscript{2}},
 \textbf{Sukairaj Hafiz Imam\textsuperscript{4}},
 \textbf{Alemu Belay Tessema\textsuperscript{1}},
 \\
 \textbf{Mohammed Oumer Adem\textsuperscript{1}},
 \textbf{Tadesse Amare Belay\textsuperscript{1}},
 \textbf{Robert Geislinger\textsuperscript{3}},
 \\
 \textbf{Umma Aliyu Musa\textsuperscript{3}},
 \textbf{Martin Semmann\textsuperscript{3}},
 \textbf{Shamsuddeen Hassan Muhammad\textsuperscript{4}},
\\
 \textbf{Henning Schreiber\textsuperscript{3}},
 \textbf{Seid Muhie Yimam\textsuperscript{3}},
\\
\\
 \textsuperscript{1}Wollo University,
 \textsuperscript{2}Bahir Dar University,
 \textsuperscript{3}Universität Hamburg,
 \textsuperscript{4}Bayero University, Kano,
\\
 \small{
   \{\href{mailto:userdavek@gmail.com}{userdavek@gmail.com}, \href{mailto:bedruy4@gmail.com}{bedruy4@gmail.com}, \href{mailto:tadesseit@gmail.com}{tadesseit@gmail.com}, \href{mailto:seidymam@gmail.com }{seidymam@gmail.com }
 \}
 }
}

\begin{document}
\maketitle
\begin{abstract}
This work explores fine-tuning OpenAI's Whisper automatic speech recognition (ASR) model for Amharic, a low-resource language, to improve transcription accuracy. While the foundational Whisper model struggles with Amharic due to limited representation in its training data, we fine-tune it using datasets like Mozilla Common Voice, FLEURS, and the BDU-speech dataset. The best-performing model, Whisper-small-am, significantly improves when fine-tuned on a mix of existing FLEURS data and new, unseen Amharic datasets. Training solely on new data leads to poor performance, but combining it with FLEURS data reinforces the model, enabling better specialization in Amharic. We also demonstrate that normalizing Amharic homophones significantly enhances Word Error Rate (WER) and Bilingual Evaluation Understudy (BLEU) scores. This study underscores the importance of fine-tuning strategies and dataset composition for improving ASR in low-resource languages, providing insights for future Amharic speech recognition research.
\end{abstract}

\section{Introduction}
Speech is one of the most fundamental and natural forms of human communication, enabling the exchange of ideas, emotions, and information across people, cultures, and generations. The transcription of speech into text,  a process known as \textit{speech-to-text (STT)}, has evolved significantly over time. Historically, this task was performed by humans, such as stenographers or transcribers, who manually convert the spoken language into written form. However, computational technologies, which are machine-based STT systems, have emerged and are revolutionizing the field. Early attempts at Automated Speech Recognition (ASR) in the mid-20th century had relied on rule-based systems and limited vocabularies \citep{jurafsky2000speech}. Over decades, advancement in machine learning, particularly deep learning, has enabled the development of more accurate and robust STT systems than rule-based systems, capable of handling diverse languages, accents, and domains \citep{hinton2012deep}.

ASR is a machine learning technology designed to convert spoken language into written text, facilitating seamless communication between humans and machines \cite{saksamudre2015review,kheddar2024automatic}. This technology has become a cornerstone of modern voice-driven systems, enabling a wide range of applications across various sectors. In healthcare, ASR is used for transcribing medical records and assisting in diagnostics, while in marketing, it powers voice-activated customer service tools and personalized advertising. In education, ASR supports language learning and accessibility tools for students with disabilities. Additionally, ASR plays a critical role in cultural preservation by transcribing and archiving oral histories and in military applications for real-time communication and command systems \cite{kumar2024comprehensive,yin2024exploration}. The versatility of ASR underscores its importance in bridging the gap between spoken language and digital technology, making it an indispensable tool in today’s increasingly voice-driven world.


Recent developments in multilingual ASR systems such as Whisper - an ASR system developed by OpenAI\footnote{\url{https://openai.com/index/whisper/}} trained on 680K hours of multilingual and multitask supervised data collected from the web, present opportunities to enhance transcription capabilities for low resource languages \cite{radford2023robust}. However, these models often face challenges when dealing with low-resource languages such as Amharic \cite{hseyin_polat__2024,yu2021finetuningpretrainedlanguagemodel}. Various strategies have been proposed to optimize Whisper's performance \cite{kummervold2024whispering, timmel2024fine, li2024improving}, achieving notable improvements in transcription accuracy and adaptability for low-resource languages other than Amharic.

Based on Whisper's improvement approaches for other low-resource languages, our study focuses on fine-tuning the Whisper model specifically for Amharic ASR and explores the capability, fine-tuning strategy, and evaluation mechanisms that lead to effective results. 
Our approach utilizes a comprehensive and diverse dataset, incorporating publicly available speech corpora such as Mozilla Common Voice \cite{mozillacv2024}, Google/FLEURS \cite{goyal2021flores101evaluationbenchmarklowresource}, and the BDU Speech Corpus \cite{assfaw2022dialect}. The main contributions of this work are three folds: 1) we extensively fine-tune and evaluate various versions of models fine-tuned on the Whisper small ASR model, and 2) we investigated the dynamic of fine-tuning with mixed datasets to reinforce the model's specialization on Amharic. 3) We investigated the impact of homophone normalization in the evaluation of the whisper-small model, which can be a good point to consider while evaluating other models and also for fine-tuning as a further exploration.

\section{Related works}
In recent years, the rise of multi-modal large language models (MM LLMs) has further expanded the capabilities of artificial intelligence to understand human language. These models, such as OpenAI's GPT-4 and Google's Gemini, integrate text, audio, and visual data to perform complex tasks, ranging from text to audiovisual understanding \cite{caffagni-etal-2024-revolution}. In the era of speech-related LLMs, Whisper by OpenAI has set a new benchmark in STT and TTS tasks as an SOTA model trained on a massive dataset of multilingual and multitask supervised data \citep{radford2023robust}. The multilingual nature of the whisper model creates the ability to generalize across low-resource languages, which makes it valuable for underrepresented linguistic communities and a best candidate to adopt its capability by fine-tuning with a more refined language-specific dataset.


~\\\textbf{STT  in Low-resource languages} Amharic is one of the languages with progress made to provide techno-linguistic tools, datasets, and research for downstream NLP tasks \cite{tonja-etal-2023-natural}. However, there is an insufficient research focus in speech-related tasks like TTS and STT along with resources, which hinders the development of accurate and reliable ASR systems. The lack of these datasets and tools limits the access to technology-driven opportunities. In the context of LLMs, low-resource languages are often underrepresented in training datasets, leading to suboptimal performance and biased outcomes. This underscores the need for fine-tuning pre-trained LLMs like Whisper to adapt them to the unique phonetic, syntactic, orthographic, and semantic characteristics of the languages \citep{Zhong2024OpportunitiesAC,hangya-etal-2022-improving}.


~\\ \textbf{Whisper and Low-Resource Languages}
OpenAI’s Whisper, a Transformer-based multilingual ASR model trained on 680k hours of diverse audio data \cite{radford2023robust}, has demonstrated state-of-the-art zero-shot performance across numerous languages. However, its efficacy diminishes for low-resource languages like Amharic, which are underrepresented in its training corpus. Recent studies highlight that fine-tuning Whisper on language-specific data mitigates this limitation. For instance, \citet{hseyin_polat__2024} applied Low-Rank Adaptation (LoRA) to optimize Whisper for Turkish, achieving parameter-efficient adaptation with minimal computational overhead. Similarly, \citet{jinpeng_li__2024} reduced the Word Error Rate (WER) for Kazakh by over 10\% through dynamic data augmentation and model quantization. \citet{shivangi_singh__2024} improved Hindi ASR performance by integrating transfer learning with Whisper’s pre-trained encoder-decoder architecture. These efforts underscore the adaptability of Whisper to linguistically diverse, low-resource settings.

Advancements in research on low resource languages have focused on addressing data scarcity and linguistic complexity. \citet{Ejigu2024} curated a foundational ASR dataset of 128 hours of Amharic speech, enabling targeted fine-tuning and data augmentation. Building on this, multilingual acoustic modeling approaches leveraging phonetically related languages reduced WER from 23.23\% to 21.52\% \cite{solomon_teferra__2024}. Furthermore, \citet{samuael_adnew__2024} introduced a transformer-based post-processing framework to refine ASR outputs, achieving a Character Error Rate (CER) of 5.5\% and WER of 23.3\% by enforcing grammatical and semantic coherence. Despite these efforts, Whisper’s adaptation to Amharic is underexplored. This work evaluates one of Whisper's multi-lingual models for the Amharic language by compiling available Datasets from sources such as the Amharic Speech Corpus \cite{Ejigu2024}, and the Amharic dataset from Google's  FLEURS and Mozilla's Common Voice.

\section{Data Collection and Preparation}
\subsection{Data Sources:}
For our study, we used publicly available Amharic ASR datasets, including BDU-speech data \cite{Ejigu2024}\footnote{\url{https://figshare.com/articles/dataset/Yohannes\_A\_Ejigu\_Amharic\_ASR\_Dataset\_zip/24959727}}, the Amharic dataset from Mozilla Common Voice\footnote{\url{https://huggingface.co/datasets/mozilla-foundation/common\_voice\_17\_0}} \cite{mozillacv2024}, and Google FLEURS ddataset\footnote{\url{https://huggingface.co/datasets/google/fleurs}}\citep{fleurs2022arxiv}.

\subsubsection{Mozilla Common Voice Data}

For this study, we used the Amharic data from version 17.0 of Mozilla's Common Voice dataset \cite{mozillacv2024}, a multilingual speech corpus designed for ASR research. This open-source, community-driven dataset includes over 31,175 hours of recorded speech, with 20,408 hours validated across 124 languages. It features demographic metadata like age, gender, and accent, which enhance speech recognition accuracy.

Data is collected through a participatory model, where volunteers read sentences or validate recordings. Each audio clip has a corresponding text transcription and undergoes peer review to ensure quality. The Amharic subset, detailed in Table \ref{tab:amharicdataset}, provides a valuable resource for improving ASR in low-resource languages.

By leveraging Common Voice, we aim to enhance Whisper's transcription capabilities for Amharic. Its open-access nature, under a Creative Commons CC0 license, supports our goal of advancing ASR technology for underrepresented languages.

\subsubsection{FLEURS Data}

Few-shot Learning Evaluation of Universal Representations of Speech (FLEURS) \cite{fleurs2022arxiv} is a benchmark designed for low-resource languages, offering an n-way parallel speech dataset across 102 languages. Built on the FLoRes-101 machine translation benchmark, FLEURS provides approximately 12 hours of speech data per language, aligned with text, making it valuable for tasks like Automatic Speech Recognition (ASR) and Speech Language Identification.

The Amharic portion of FLEURS includes high-quality transcriptions, which are crucial for improving ASR accuracy. Although the base Whisper model was trained on FLEURS, Amharic was overshadowed by higher-resource languages, leading to suboptimal performance. To address this, we fine-tuned Whisper specifically on the Amharic data from FLEURS, significantly improving transcription accuracy.

FLEURS' natural speech recordings and robust quality control ensure that fine-tuning produces a model well-suited to Amharic's linguistic intricacies. This focus enhances the usability and performance of speech technologies for underrepresented languages like Amharic.

\subsubsection{Bahir Dar University Noisy Amharic Speech Dataset (BDU-speech)}

In our study, we utilized the \textit{BDU Speech Corpus - Bahir Dar University Noisy Amharic Speech Dataset} \citep{Ejigu2024}, which plays a crucial role in evaluating ASR models under realistic noisy conditions. This dataset includes audio recordings from the Sidama region, featuring 400 sentences spoken by 50 individuals, amounting to 20k sentences with a total duration of approximately 44 hours and 46 minutes. Designed to simulate real-world challenges, the audio clips vary from 4 to 20 seconds and include significant background noise, speech distortions, and non-native accents. Each recording is properly transcribed, providing a reliable ground truth for training and evaluation. The audio data is processed into spectrograms using Short-Time Fourier Transform (STFT) to ensure compatibility with deep learning architectures like CNNs and RNNs. This dataset has been instrumental in testing the Whisper model's robustness and transcription accuracy in challenging environments.

\begin{table}[]
\centering
\begin{tabular}{lll}
\toprule
\textbf{Dataset Source} &\multicolumn{2}{l}{\textbf{\# of instances}}\\
\hline
& \textbf{Train} & \textbf{Test} \\
\hline
BDU Speech Corpus    & 10,875 & 389 \\ 
Mozilla Common Voice & 698 & 205   \\ 
FLEURS    & 3609 & 516 \\  
\bottomrule
\end{tabular}
\caption{ Speech data sources.}
\label{tab:amharicdataset}

\end{table}



\subsection{Audio Preprocessing:}
Proper audio processing is a foundational step in fine-tuning Whisper for Amharic datasets. Resampling audio to 16 Khz ensures compatibility with the model, and tokenization prepares the transcriptions to be ready for the model to understand, along with the Whisper feature extractor to extract audio features to make them ready for training, and test sets enable effective model evaluation. These steps collectively ensure that the Whisper model can be fine-tuned to achieve optimal performance on Amharic speech recognition tasks.

\section{Experimentation}
\subsection{Fine-Tuning Process}
We fine-tuned the \textbf{Whisper-small} model, a smaller version of OpenAI's Whisper architecture, using several Amharic datasets. The fine-tuning process involved the following steps:
\textbf{Data Preparation}: Datasets from their source are fetched and resampled to 16khz, which is the sampling rate compatible with the whisper fine-tuning.\\
\textbf{Feature Extraction}: Both audio and text features get extracted by the Whisper feature extractor for Audio and the Whisper tokenizer for text.\\
\textbf{Pretrained Model}: We started with the pre-trained Whisper-small model, which was initially trained on a large multilingual corpus.

\begin{figure*}[!ht]
    \centering
    \includegraphics[width=\linewidth]{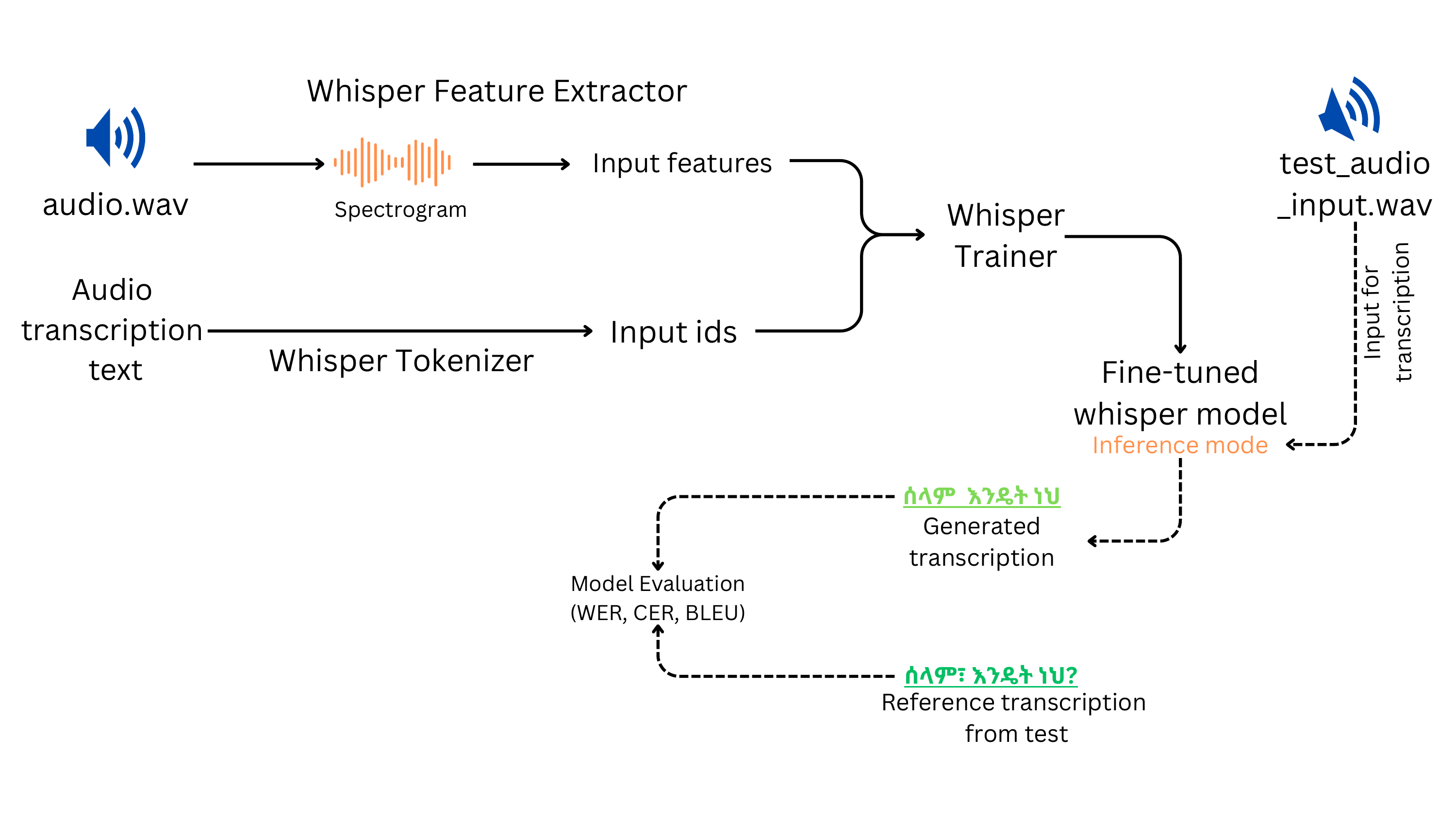}
    \caption{The fine-tuning process}
    \label{fig:finetuning}
\end{figure*}

\textbf{Training Configuration}: The model was fine-tuned with the following hyperparameters:
\begin{itemize}
    \item Batch Size: 16
    \item Learning Rate: 5e-5
    \item Epochs: 5
    \item Max Generation Length: 256
    \item Hardware: The finetuning of the Whisper model utilizes a GPU infrastructure composed of NVIDIA A100 80GB PCIe and NVIDIA H100 NVL GPUs, providing substantial memory and computing power to handle large-scale data processing efficiently.
    \item Optimization: we use half-precision floating point (FP16) for optimizing the learning model to speed up training and reduce memory usage.
\end{itemize}
The complete process of fine-tuning is shown in Figure \ref{fig:finetuning}

The fine-tuning was performed on the following datasets:
\begin{itemize}
    \item \textbf{FLEURS Amharic Data}: The model was fine-tuned on the Amharic portion of the FLEURS dataset, which was also used in the pretraining of the Whisper model.
    \item \textbf{Mozilla Common Voice v17.0 Amharic Set}: This dataset was used to fine-tune the model on a more diverse set of Amharic speech data.
    \item \textbf{BDU Speech Corpus}: We fine-tuned the model using this Amharic Speech Corpus, which includes a variety of speech samples from different speakers.
    \item \textbf{Combined Datasets}: We also fine-tuned the model on a combination of all three datasets (FLEURS, Common Voice, and BDU speech corpus) to evaluate the impact of mixed data on model performance.
\end{itemize}

\subsection{Evaluation}
Evaluating speech-to-text (STT) models is crucial to asses the performance of the models and ensure whether they meet the desired accuracy and usability standards or not. Several metrics are commonly used to evaluate STT systems, each metric providing insights into different aspects of the model's performance. In STT, the most widely used metrics, which are employed in this work, are \textit{Word Error Rate (WER), Character Error Rate (CER), and Bilingual Evaluation Understudy (BLEU)}. Additonally, we highlight the importance of \textit{Human Evaluation} in addressing the litigation beyond these automated metrics. Word Error Rate measures the percentage of word-level errors in transcribed text compared to the reference text; a lower WER indicates better performance. Also, CER measures the percentage of character-level errors in the transcribed text, and the same applies to CER; the lower the percentage, the better the performance. BLEU measures the overlap between the model-generated text and the reference text using n-gram precision. 
Corpus BLEU and Average BLEU are used in our model evaluation to measure the overlap over the entire dataset (overall quality of the transcription) and averaged sentence level measure (for consistency across individual examples). In this work, average BLEU is used even if it is not the standard metric in STT tasks since it can provide 

\subsection{Homophone Normalization Effects} 

In Amharic writing, there are different characters with the same sound, and they are called homophones. The homophones comprise the ha sounds <\selectlanguage{ethiop}ha\selectlanguage{english} ha>,
<\selectlanguage{ethiop}.ha\selectlanguage{english} ha>, and <\selectlanguage{ethiop}_ha\selectlanguage{english} ha>, the a sounds <\selectlanguage{ethiop}'a\selectlanguage{english} 
a> and <\selectlanguage{ethiop}a\selectlanguage{english} a>, the sa sounds <\selectlanguage{ethiop}'sa\selectlanguage{english} sa> and 
<\selectlanguage{ethiop}sa\selectlanguage{english} sa>, and the sa sounds <\selectlanguage{ethiop}.sa\selectlanguage{english} sa> and <\selectlanguage{ethiop}.ca\selectlanguage{english} sa> with all including their seven consonant-vowel combinations. These characters might affect the speech recognition task positively or negatively during evaluation. For example, the impacts of such normalization effects are explored, such as Machine translation \cite{belay2021impacts}.  However, the impact of such homophone characters on SST has not yet been investigated. One of the approaches to handling these homophones in other NLP tasks is normalizing them into a single representation. In this task, we apply normalization during evaluation.

\subsection{Experimental Results}

\begin{table*}[h!]
    \centering
    \begin{tabular}{lccccc}
        \toprule
        \textbf{Models}& WER(\%) & CER(\%) & corpusBLEU(\%) & avg.BLEU\\
        \midrule
        \multicolumn{4}{l}{\textit{Fine-tuning}} \\      
        Whisper-small-am &\textbf{31.71}  & \textbf{10.18}  &  \textbf{45.66} & \textbf{43.65} \\
        whisper-small-fc-am & 47.97  & 16.60  & 29.89 & 28.06 \\
        whisper-small-am-fleurs & 39.05  & 12.71  & 40.77 & 39.12\\
        whisper-small-am-common-speech &103.17  & 81.47  & 0.014 & 0.44 \\
        whisper-small-am-v2 & 99.6 & 94.55  &  0.003 & 0.042\\
        whisper-small-am-on-aggregated & 33.44 & 11.25 & 44.28 & 42.44 \\
        \midrule
        \multicolumn{4}{l}{\textit{Evaluation on normalized references and predictions} }  \\
        Whisper-small-am &\textbf{29.19}  & \textbf{9.44}  & \textbf{49.88} & \textbf{47.89} \\
        whisper-small-fc-am &45.75  & 15.77  & 32.97 & 31.14 \\
        whisper-small-am-fleurs &36.67  & 11.97  & 44.59 & 43.03 \\
        whisper-small-am-common-speech &103  & 81.4  & 0.014 & 0.448 \\
        whisper-small-am-v2 &99.59  & 94.5  & 0.003 & 0.04 \\
        whisper-small-am-on-aggregated &30.95  & 10.56  & 48.79 & 46.86 \\
    \bottomrule
    \end{tabular}
    \caption{Evaluation results on fine-tuned version of Whisper and zero-shot on test set 1 (\textit{FLEURS's test set})}
    \label{tab:eval1}
\end{table*}

\begin{table*}[h!]
    \centering
    \begin{tabular}{lccccc}
        \toprule
        \textbf{Models}& WER(\%) & CER(\%) & corpusBLEU(\%) & avg.BLEU\\
        \midrule
        \multicolumn{4}{l}{\textit{Fine-tuning}} \\      
        Whisper-small-am &\textbf{75.75}  & \textbf{20.23}  & \textbf{45.99} & \textbf{7.76} \\
        whisper-small-fc-am & 80.46  & 23.34  & 3.01 & 6.05 \\
        whisper-small-am-fleurs & 77.94  & 22.80  & 3.77 & 6.96\\
        whisper-small-am-common-speech &99.96  & 79.71  & 0.01 & 0.043 \\
        whisper-small-am-v2 & 96.80 & 93.40  &  0.013 & 0.53\\
        whisper-small-am-on-aggregated & 108.8 & 94.0 & 0.109 & 2.07 \\
        \midrule
        \multicolumn{4}{l}{\textit{Evaluation on normalized references and predictions} }  \\
        Whisper-small-am &\textbf{74.33}  & \textbf{19.48}  & \textbf{5.61} & \textbf{8.50} \\
        whisper-small-fc-am &79.04  & 22.51  & 3.72 & 6.62 \\
        whisper-small-am-fleurs &76.99  & 22.23  & 4.33 & 7.42 \\
        whisper-small-am-common-speech &99.96  & 79.70  & 0.010 & 0.04 \\
        whisper-small-am-v2 &96.80  & 93.36  & 0.01 & 0.53 \\
        whisper-small-am-on-aggregated &108.8  & 93.98  & 0.10 & 2.07 \\
    \bottomrule
    \end{tabular}
    \caption{Evaluation results on fine-tuned version of Whisper and zero-shot on test set 2 (\textit{BDU Speech data test set})}
    \label{tab:eval2}
\end{table*}

In this section, we detail the experiments conducted to fine-tune the Whisper model for Amharic speech recognition. Our goal is to evaluate the performance of various fine-tuned versions of the Whisper model using different datasets and configurations. We also explore the impact of homophone normalization on the evaluation metrics. The evaluation test sets are: 
\begin{enumerate}
    \item FLEURS Test Set: 516 samples
    \item BDU Speech Data Test Set: 389 samples
    \item Common Voice Test Set: 205 samples
\end{enumerate}
The results are summarized in Tables \ref{tab:eval1}, \ref{tab:eval2}, and \ref{tab:eval3}, which show the performance of each model on the respective test sets. The Whisper-small-am model, fine-tuned on the combination of FLEURS and Common voice data, consistently performed the best across all test sets, with significant improvements in WER, CER, and BLEU scores after homophone normalization. details on which model trained on which data is explained in Appendix 
 \ref{sec:appendix1}.

Models were evaluated on three different test sets from those training datasets. So, we have 3 evaluation results for each test set. Common voice test set: 205; fleurs test set: 516; BDU Speech corpus: 359
\begin{table*}[!ht]
    \centering
    \begin{tabular}{lccccc}
        \toprule
        \textbf{Models}& WER(\%) & CER(\%) & corpusBLEU(\%) & avg.BLEU\\
        \midrule
        \multicolumn{4}{l}{\textit{Fine-tuning}} \\      
        Whisper-small-am &\textbf{59.4}  & \textbf{23.03}  & \textbf{20.83} & \textbf{22.92} \\
        whisper-small-fc-am &62.67  & 23.22  & 16.93 & 19.53 \\
        whisper-small-am-fleurs &71.95  & 28.40  & 11.89 & 14.41\\
        whisper-small-am-common-speech &113  & 84.6  & 0.035 & 0.16 \\
        whisper-small-am-v2 & 97.62 & 90.9  & 0.17 & 0.97\\
        whisper-small-am-on-aggregated & 60.2 & 24.46 & 20.46 & 21.53 \\
        \midrule
        \multicolumn{4}{l}{\textit{Evaluation on normalized references and predictions} }  \\
        Whisper-small-am &\textbf{57.98}  & \textbf{22.42}  & \textbf{22.27} & \textbf{24.17} \\
        whisper-small-fc-am &61.46  & 22.61  & 18.38 & 20.80 \\
        whisper-small-am-fleurs &70.74  & 27.69  & 12.79 & 15.33\\
        whisper-small-am-common-speech &113  & 84.53  & 0.035 & 0.16 \\
        whisper-small-am-v2 & 97.62 & 90.73  & 0.17 & 0.97\\
        whisper-small-am-on-aggregated & 58.77 & 23.88 & 21.82 & 22.83 \\
    \bottomrule
    \end{tabular}
    \caption{Evaluation results on fine-tuned version of Whisper on test set 3 (\textit{common voice's test set})}
    \label{tab:eval3}
\end{table*}

\subsection{zero-shot learning}
We also conducted a zero-shot learning experiment using the pre-trained Whisper-small, Whisper-medium, and Whisper-Large models. In these experiments, the models were tested in a small sub-set of test sets without any fine-tuning. The results were poor, with the models generating non-Amharic characters, other language texts, gibberish content, and on the Whisper-large v3 and Whisper-large v3-turbo, repetitive Amharic letters. Also, the evaluation metrics results are extremely poor. This highlights the need for fine-tuning the whisper model specifically for Amharic, as the pre-trained models struggle to generalize to low-resource languages like Amharic without additional training. 

\textbf{Normalization Effects}: 
In Amharic, the presence of homophone characters can lead to variations in model predictions, where the generated output may use different but phonetically similar characters compared to the reference sentences. To assess the impact of these variations on evaluation metrics, normalization was applied to both the reference and predicted sentences after the model generated its outputs. This process aimed to minimize the discrepancies caused by homophone character differences and provide a more accurate evaluation of the model's performance.

As shown in the tables of evaluation; the results demonstrate that applying normalization significantly impacts the evaluation metrics across the fine-tuned Whisper models. Notably, after normalization, almost all models have shown improvements in the evaluation metrics used in this work. For instance, in Table \ref{tab:eval1}, \textit{whisper-small-am} has improved its WER from 31.71\% to 29.19\%, and the BLEU score increased from 45.66\% to 49.88\%.  These improvements highlight how homophone variations can distort the evaluation metric, and normalization helps mitigate such issues by aligning phonetically similar but orthographically different outputs. This trend indicates that normalization has been a crucial step for fair and accurate model assessment in Amharic, where homophone characters are prevalent. However, the normalization effect on the training data is unexplored and needs further investigation, and we look forward to exploring it.

\textbf{Human Evaluation}: In addition to evaluation using various metrics, the model's output was observed and analyzed on a separate test set to assess its performance on the given data. Based on this analysis, the top three performing models demonstrated strong results in transcribing audio files as well as in handling direct recordings through the Gradio inference interface. While automated metrics like WER, CER, and BLEU provide quantitative measures of performance, they have limitations on semantic meaning and contextual correctness. Also, the transcribed text's fluency, naturalness, or usability can't be measured by those metrics unless by human evaluation.

To further enhance the output, additional post-processing tasks are needed to be applied to the model's predictions using Amharic-specific tools such as Named Entity Recognition (NER), spell checkers, and grammar checkers. These tools help correct missed or incorrectly transcribed characters, leading to improved results. However, this post-processing step requires further investigation to optimize its effectiveness.




\section{Discussion}
The experiments demonstrate that fine-tuning the Whisper model on Amharic-specific data significantly improves its performance, especially when the model is trained on a combination of existing (FLEURS) and new (Common Voice, BDU) datasets. The model's ability to generalize to unseen data improves when it is exposed to a diverse set of speech samples, including noisy and dialect-heavy recordings. However, fine-tuning on only new, unseen data without reinforcement from existing data leads to suboptimal performance, as the model struggles to adapt to the new linguistic patterns.

In conclusion, the experiments highlight the importance of dataset composition and fine-tuning strategies for improving ASR performance in low-resource languages like Amharic. Future work could explore the impact of continual learning and data augmentation techniques to further enhance the model's robustness and accuracy.



\section*{Limitations}
While this study provides valuable insights into the fine-tuning of the Whisper model for Amharic speech recognition, it has several limitations that should be acknowledged. These limitations highlight areas for future research and improvement.
\begin{itemize}
    \item \textbf{Limited to Whisper-small Model}: The experiments were conducted only on the Whisper-small model, which may not fully capture the potential of larger variants like Whisper-medium or Whisper-large. Future work should explore fine-tuning these larger models for better performance.
    \item \textbf{No Comparison with Other Multilingual Speech LLMs}: The study does not compare Whisper with other multilingual speech LLMs (e.g., SeamlessM4T, Google's USM). Future research should include such comparisons to identify the most effective model for Amharic ASR.
    \item \textbf{Focus on Fine-Tuning Strategies:} The work primarily explores fine-tuning strategies and evaluation factors but does not extend to advanced techniques like continual learning, data augmentation, or transfer learning. These could further enhance model performance.
    \item \textbf{Dataset Limitations}: The datasets used, while diverse, are relatively small. Expanding to include more speakers, dialects, and noisy conditions would improve the model's robustness and generalization capability.
    \item \textbf{Homophone Normalization on Training Data}: Homophone normalization was applied only during evaluation. Its impact on the training process remains unexplored and could potentially improve the model's handling of Amharic's orthographic challenges.
    \item \textbf{Post-Processing Tools}: The study highlights the potential of post-processing tools like NER, spell checkers, and grammar checkers but does not integrate them. Incorporating these tools could further improve transcription accuracy.
\end{itemize}


\bibliography{acl_latex}

\appendix

\section{Fine-tuned model variants}
\label{sec:appendix1}
\textbf{Model variants trained on different datasets}
\begin{itemize}
\item whisper-small-am-fleurs - It is trained on FLEURS Amharic data [FLEURS - Amharic portion], which is used in the pre-training of the Whisper model.
\item whisper-small-am-common-speech - This version is trained on Mozilla's common voice v17.0 Amharic set
\item pre-trainedwhisper-small-am-v2 - Trained using the Amharic speech corpus dataset

\item whisper-small-fc-am - By using common voice data, this model is trained on the FLEURs fine-tuned version model \textit{whisper-small-am-fleurs}

\item Whisper-small-am - trained on the combined data from FLEURs and common voice's train set

\item whisper-small-am-on-aggregated - Trained on the combination of all 3 data sources [FLEURS + Common voice 17.0 + Amharic speech corpus]
\end{itemize}

\end{document}